## Title

# ***In silico*** **Deep Learning Protocols for Label-Free Super-Resolution Microscopy: A Comparative Study of Network Architectures and SNR Dependence**

## List of authors

Shiraz S. Kaderuppan[1,2,*], Jonathan D. Mar[2,3], Andrew Irvine[4], Anurag Sharma[1,2], Muhammad Ramadan Saifuddin[1,2], Wai Leong Eugene Wong[5], Wai Lok Woo[6]

## Affiliations

1. Newcastle University in Singapore, 1 Punggol Coast Road, Level 2, Block E1, Singapore 828608
2. Faculty of Science, Agriculture and Engineering (SAgE), Newcastle University, Newcastle upon Tyne NE1 7RU, U.K.
3. Department of Electrical, Computer, and Biomedical Engineering, Toronto Metropolitan University, Toronto, Ontario, M5B 2K3, Canada
4. The Cavendish Laboratory, Department of Physics, JJ Thomson Avenue, University of Cambridge, Cambridge CB3 0HE, UK
5. Engineering Cluster, Singapore Institute of Technology, 10 Dover Drive, Singapore 138683
6. Computer and Information Sciences, Northumbria University, Newcastle upon Tyne, NE1 8ST, UK

**Corresponding author*: Shiraz S. Kaderuppan (S.S.O.Kaderuppan@newcastle.ac.uk)

## Abstract

The field of optical microscopy spans across numerous industries and research domains, ranging from education to healthcare, quality inspection and analysis. Nonetheless, a key limitation often cited by optical microscopists refers to the limit of its lateral resolution (typically defined as ~200nm), with potential circumventions involving either costly external modules (e.g. confocal scan heads, etc) and/or specialized techniques [e.g. super-resolution (SR) fluorescent microscopy approaches]. Addressing these challenges in a normal (non-specialist) context thus remains an aspect outside the scope of most microscope users & facilities. This study thus seeks to evaluate an alternative & economical approach to achieving SR optical microscopy, involving *non-fluorescent* phase-modulated microscopical modalities such as Zernike phase contrast (PCM) and differential interference contrast (DIC) microscopy. Two *in silico* deep neural network (DNN) architectures which we developed previously (termed O-Net and Θ-Net) are assessed on their abilities to resolve a custom-fabricated test target containing nanoscale features calibrated via atomic force microscopy (AFM). The results of our study demonstrate that although both O-Net and Θ-Net seemingly performed well when super-resolving these images, they were *complementary* (rather than *competing*) approaches to be considered for image SR, particularly under different image signal-to-noise ratios (SNRs). High image SNRs favoured the application of O-Net models, while low SNRs inclined preferentially towards Θ-Net models. These findings demonstrate the importance of model architectures (in conjunction with the source image SNR) on model performance and the SR quality of the generated images where DNN models are utilized for non-fluorescent optical nanoscopy, even where the same training dataset & number of epochs are being used.

## Introduction

Super-resolution optical microscopy (SROM) has provided an avenue for the conduct of nanoscopical studies without the need for high accelerating voltages, vacuum desiccation [1], ultramicrotomy [2] and heavy metal staining [3] and/or cryogenic cooling [4] as is often utilized in electron microscopy techniques or through contact-based approaches such as cantilever probes in atomic force microscopy (AFM) [5] and scanning probe microscopy (SPM) [6]. Fundamentally, super-resolution microscopy derives its nomenclature from the recent advent of overcoming the Rayleigh diffraction limit imposed by the optical microscope – a barrier which remained unsurpassed for over a century since it was first proposed in 1896 [7] and which is described by Equation (1) as follows [8]:

Rayleigh Criterion, $d_{x,y} = \frac{0.61\lambda}{NA_{obj}}$ (for fluorescence) ***or*** $\frac{1.22\lambda}{NA_{cond}+NA_{obj}}$ (for brightfield) ----- (1)

Intriguingly, SROM has manifested in serval flavours over the decades, including (amongst others) confocal laser scanning microscopy (CLSM) [9], 4Pi [10], and NSOM [11]. However, the concept of SROM only gained traction with the introduction of *true* fluorescence optical nanoscopy techniques breaching the 100nm lateral resolution limit – a feat realized through the development of Stimulated Emission Depletion (STED) microscopy [12] and single molecule localization microscopy (SMLM) approaches (such as PALM [13] and STORM [14], amongst others). Recently, a variant of STED, named MINSTED, has allowed the attainment of lateral resolutions down to $\sigma = 4.7$ Å [15], proving that optical nanoscopy has the capability of surpassing the Rayleigh diffraction limit by up to 3 orders of magnitude and attaining resolutions transcending even that of scanning electron microscopy (SEM) approaches – the Verios 5 XHR SEM attaining a lateral resolution down to 0.6nm at 30kV for comparison [16]. In this respect, it would be noteworthy to heed the relative growing prominence of optical microscopy in the nanoscale domain (i.e. optical nanoscopy).

Nonetheless, a major caveat plagues the current portfolio of optical nanoscopy techniques as highlighted previously. These approaches are primarily *fluorescent* microscopy modalities which rely on the localization of fluorophores to detect and identify the tagged structures of interest. These structures may include specific proteins both in and out of the cell, or attached to the cellular membrane/envelope of specific organelles/viruses. Consequently, common issues which may impede the accuracy of fluorescence localization approaches (including photobleaching, *in vivo* environmental disruptions caused by the extraneous introduction of fluorophores, cross-talk, non-specific binding of fluorophores to the target protein/structure of interest, etc) are amongst the problems encountered for these approaches. In this regard, *label-free* fluorescence (often through an approach known as *multiphoton microscopy*) has been proposed as an alternative although the fluorescence excitation characteristics of native molecules/metabolites which might be utilized for this purpose is generally limited to a few characterized markers (e.g. NADH [17]). Hence an impending need exists to fill this void – a niche domain known as *label-free optical nanoscopy*. In this respect, some computational approaches have been developed (such as ANNA-PALM [18] & Deep-STORM [19]), while these utilize the corresponding fluorescence nanoscopical image as the designated *ground truth* for the training of deep neural network (DNN) models to super-resolve the input *widefield epifluorescent* micrographs. Separately, other studies (such as [20], [21]) have also developed DNN models aimed at super-resolving micrographs acquired via popular *non-fluorescent* optical microscopy approaches, namely phase contrast microscopy (PCM) & differential interference contrast (DIC) microscopy. In this context, we seek to assess the SR performance of the O-Net and Θ-Net models which we developed using a specially prepared custom-fabricated and calibrated test target sample, containing features spanning from 20nm – 1µm. The methodology employed in this study (coupled with the results of this analysis) are presented in the subsequent sections of this paper.

## Methodology

### *Image acquisition for model training datasets*

As was described in our previously-published studies ([20], [21]), identical model training image datasets were used in the present context, these comprising of PCM (Zernike) & DIC micrographs acquired from a large variety of different samples (plant & animal tissues, as well as protists). The hardware setup consisted of a Leica N PLAN L 20X/0.4 Corr Ph1 objective (Leica P/N: 506058) and a Leica HCX PL Fluotar L 40X/0.60 Corr Ph2 objective (Leica P/N: 506203) mounted on a Leica DM4000M microscope & having a CMOS camera (RisingCam® E3ISPM12000KPA, RisingTech) with a pixel size of 1.85µm × 1.85µm and an EK 14 Mot motorized stage (Märzhäuser Wetzlar GmbH & Co. KG). The 20X/0.40 objective was used for acquiring low-resolution (LR) images, while the 40X/0.60 objective was used to obtain high-resolution (HR) images. Similar regions of interest (ROIs) were imaged of the samples under both PCM & DIC microscopy via a self-developed desktop app coded in C# .NET, prior to being registered, cropped and split into 256x256 RGB tiles in MATLAB R2020a (© 1984-2020, The MathWorks, Inc) [21]. These image tiles were then concatenated into LR (**Source**)-HR (**Expected**/target) image pairs (3944 image pairs per imaging modality employed), which were then converted into NumPy arrays for training the individual nodes of the proposed Θ-Net network in Python 3.9 [21].

### *DNN Architecture & Model training*

For the purposes of this study, the DNN models adopting the O-Net and Θ-Net architectures were trained on a workstation equipped with an Intel® Core™ i9-10920X CPU, 128GB RAM and a NVIDIA RTX™ 3090 GPU (© 2022, NVIDIA Corporation). The model architectures used were similar to that described in [21], i.e. a single-node 5-layer/7-layer O-Net architecture (for the O-Net models) or a triple-node O-Net architecture (for the Θ-Net models), with Figure 1 depicting the details of each of these architectures:

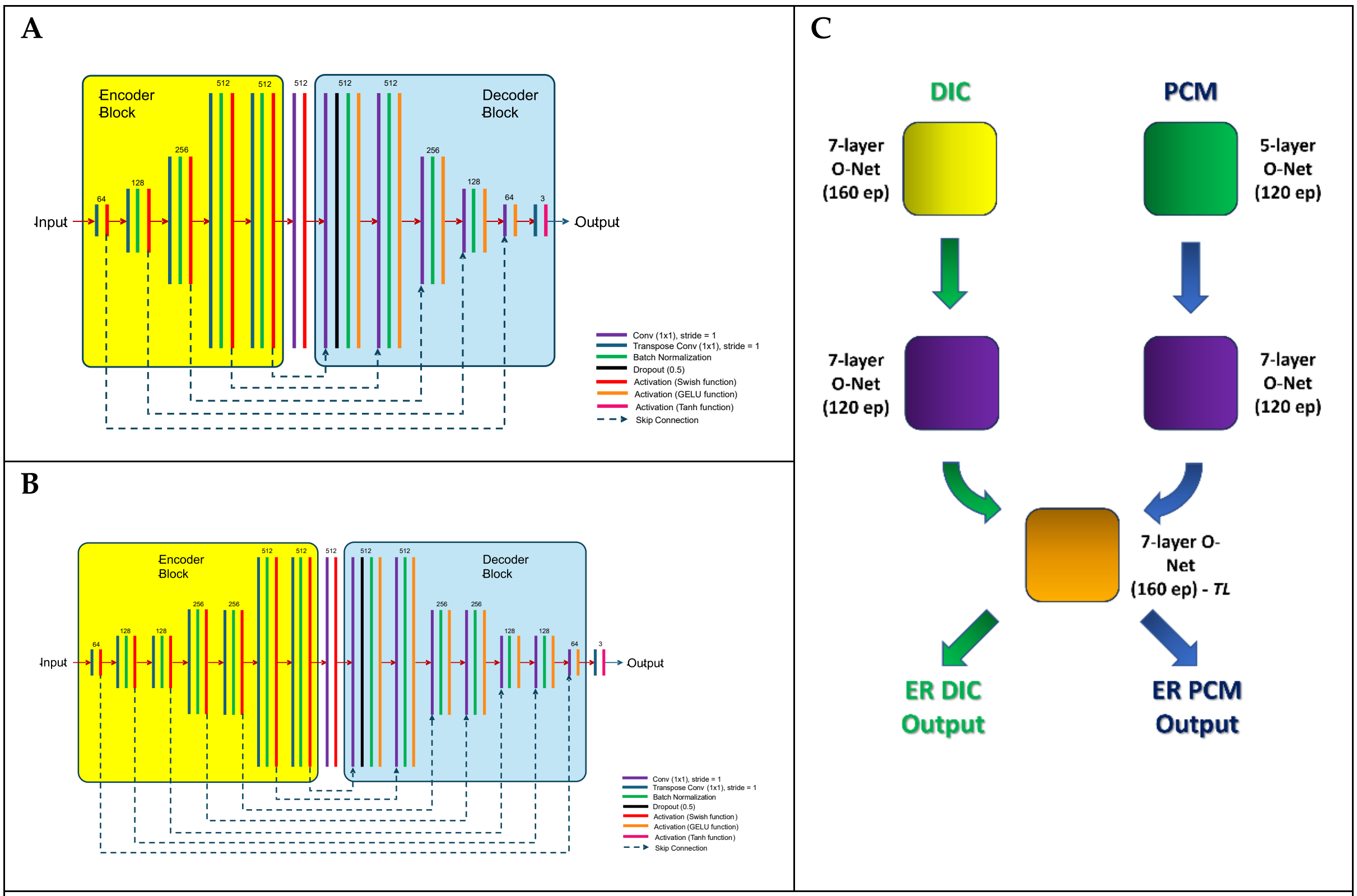

Θ **Figure 1**: O-Net DNN architectures, employing either a 5-layer (**A**) or 7-layer (**B**) network. As highlighted in [20], each O-Net architecture contains skip-connections (*concatenations*) linking transposed convolution layers (in the encoder block) to convolution layers (in the decoder block). Concatenation of multiple O-Net frameworks (*nodes*) results in the formation of a Θ-Net model framework (an example of which is depicted in **C**), as described in [21] and in the current context. In the present study, both the DIC & PCM model training pipelines utilized differing O-Net models, with the exception of the 3rd (and final) node – a common O-Net model trained on *both* the DIC & PCM image datasets (i.e. *transfer learning*). Figure Source: [21].

For the U-Net models, the model architecture as described in [20] (a 5-layer U-Net architecture) was followed.

*Validation of trained models*

The trained models were validated using a calibrated test-target pattern adopting the industry-wide USAF1951 standard of equally-spaced bars and spaces was designed using KLayout 0.28.3 (© Shiraz S Kaderuppan, Newcastle University, 2024) and custom-fabricated in PMMA using electron beam lithography (EBL) on a 24mm × 24mm square coverslip (#1.5) (© Jonathan Mar, Newcastle University, and Andrew Irvine, University of Cambridge, 2024). The said target consists of features ranging between 20nm – 6µm, making it suitable for assessing the SR performance of the proposed O-Net & Θ-Net frameworks. The coverslip was then mounted onto a standard (1” × 3”) microscope glass slide using a high refractive index mountant (Norland Optical Adhesive NOA 170, $n_d$ = 1.70 (at 589nm), Norland Products Inc), prior to image acquisition (utilizing both PCM & DIC techniques) with a Leica HCX PL APO 100X objective (Leica P/N: 506211). Different intermediate magnifications of 2.4X and 3.75X were also employed in the image acquisition, resulting in a total of 4 images (2 for each of the afore-mentioned modalities). Subsequently, specific RoIs were defined for the respective images, with these RoIs being subsequently cropped into individual image tiles in MATLAB R2022b (© 1984-2022, The MathWorks, Inc) (see Figures 2**C** & **D** below for details). For the 2.4X intermediate magnification images, the RoI corresponding to the 400-600nm features was cropped into 300px × 300px image tiles (total area: 2100px × 900px, while the RoI for features ≤200nm was cropped into either 600px × 600px image tiles (for PCM) or 700px × 700px image tiles (for DIC). For the 3.75X intermediate magnification micrographs, 256px × 256px image tiles were cropped for the respective RoIs instead. The cropped images were then combined into a single ‘expanded view’ image as depicted in Figures 3-6. To ensure proper verification of the DNN model performance (especially with regards to its potential for image SR), images of the assayed RoIs were acquired separately using AFM, where the accompanying scale bar was used for precise validation of the respective feature dimensions. This thus allowed comparison with the DNN-generated SR PCM/DIC images, enabling an unbiased determination of the image SR performance for each of the evaluated models. The SR PCM/DIC images were also converted to grayscale prior to the generation of an intensity profile plot along a linear portion of the image in ImageJ 1.52n (NIH, USA), so that the individual bar widths as determined by the FWHM in the intensity profile plot could be used to specify the minimum resolvable distance, according to the Rayleigh criterion. High levels of noise present in the grayscale images (preventing the direct determination of the respective bars and spaces from the intensity profile plot) necessitated the corresponding distances to be inferred by comparing the SR PCM/DIC image against the respective AFM image, when extrapolated onto the intensity profile plot (see Figures 7 and 8 for details).

## Results

The U-Net, O-Net and Θ-Net models as described in the previous **Methodology** section were assessed on their image SR generation performance using the calibrated test-target slide, and the results of this analyses are presented in Figure 2:

*AFM image of the test target sample*

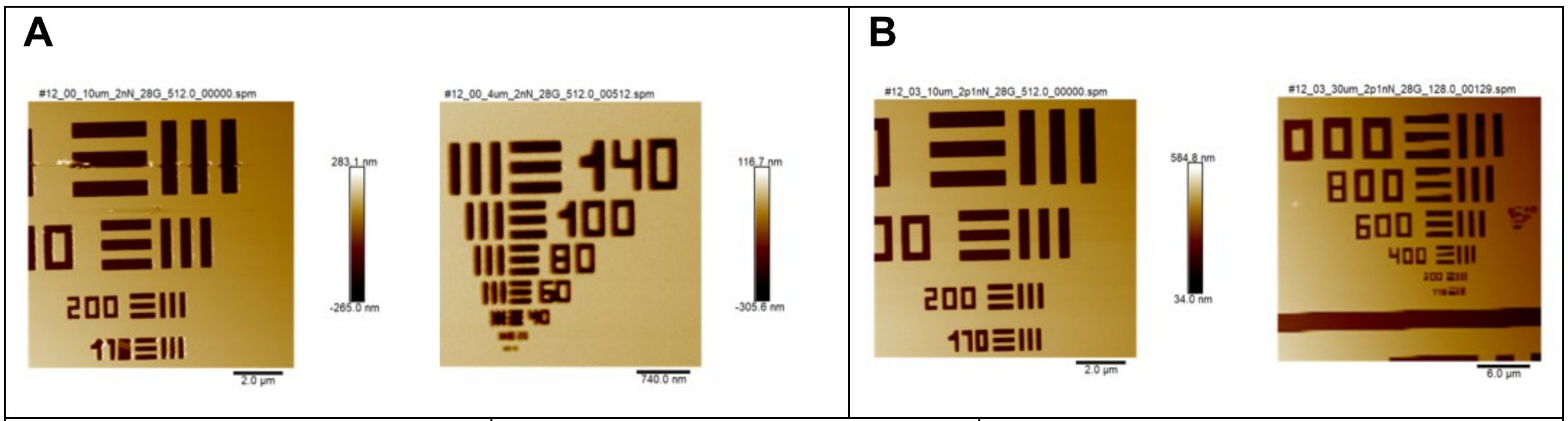

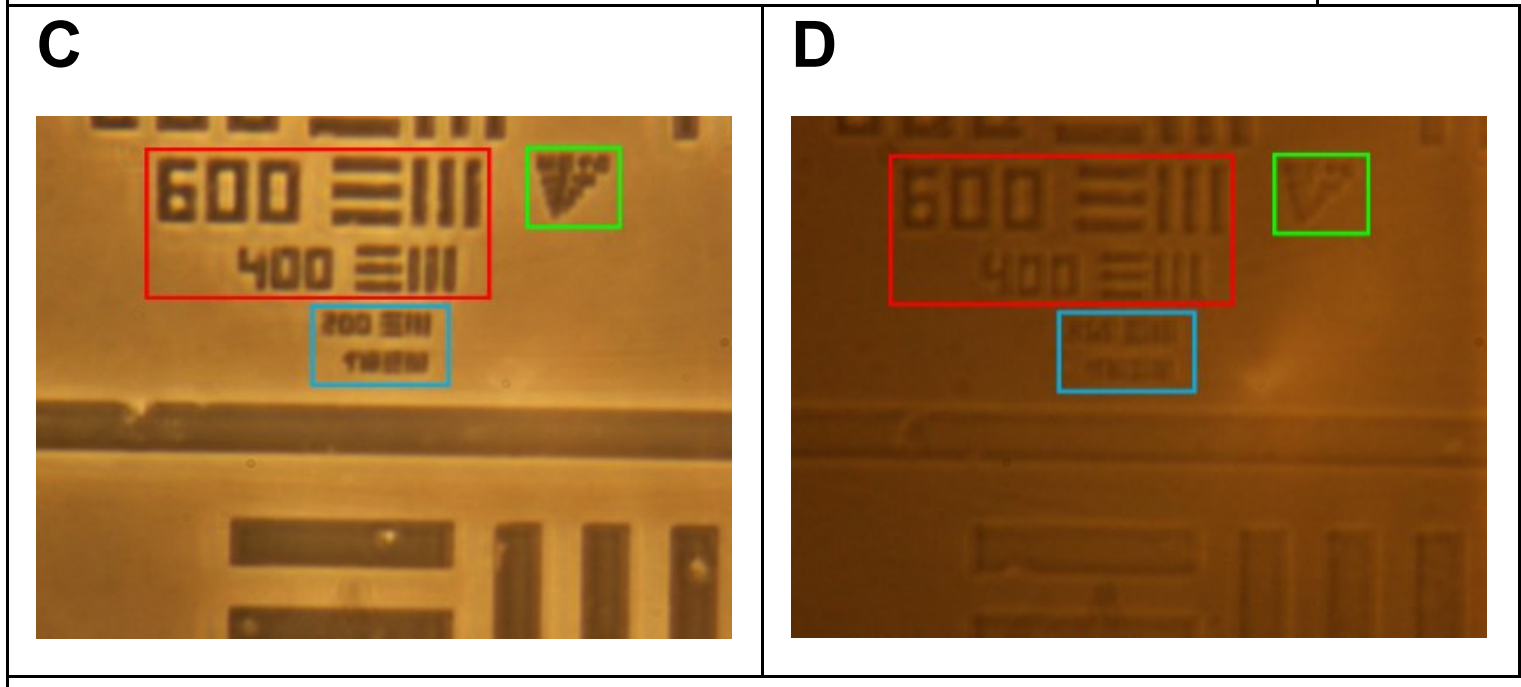

E

| λ / nm | $N.A._{obj}$ | $N.A._{cond}$ | Lateral Resolution, $d_{xy}$ (nm) |
|---|---|---|---|
| 400 | 1.4 | 0.9 | 212 |
| 550 | 1.4 | 0.9 | 292 |
| 580 | 1.4 | 0.9 | 308 |

**Figure 2**: **A** & **B** AFM images acquired of the test target pattern fabricated in PMMA (© Jonathan Mar, Newcastle University, UK and Andrew Irvine, University of Cambridge, UK). Selected RoIs are **#00** (**A**) and **#03** (**B**). **C** PCM and **D** DIC photomicrographs of these same regions (**#00** and **#03** respectively) acquired under an intermediate magnification of 2.4X. The red, blue and green boxed RoIs indicate those regions used for image SR and further analysis. **E** Table indicating the computed lateral resolutions ($d_{xy}$) based on the Rayleigh diffraction-limited equation as described in equation (1) previously [22].

Subsequently, the test target slide was imaged via PCM and DIC microscopies (described in **Methodology** previously) and subjected to *in silico* image SR via application of the aforementioned models. The following Figures 3 to 6 depict the results of this analysis:

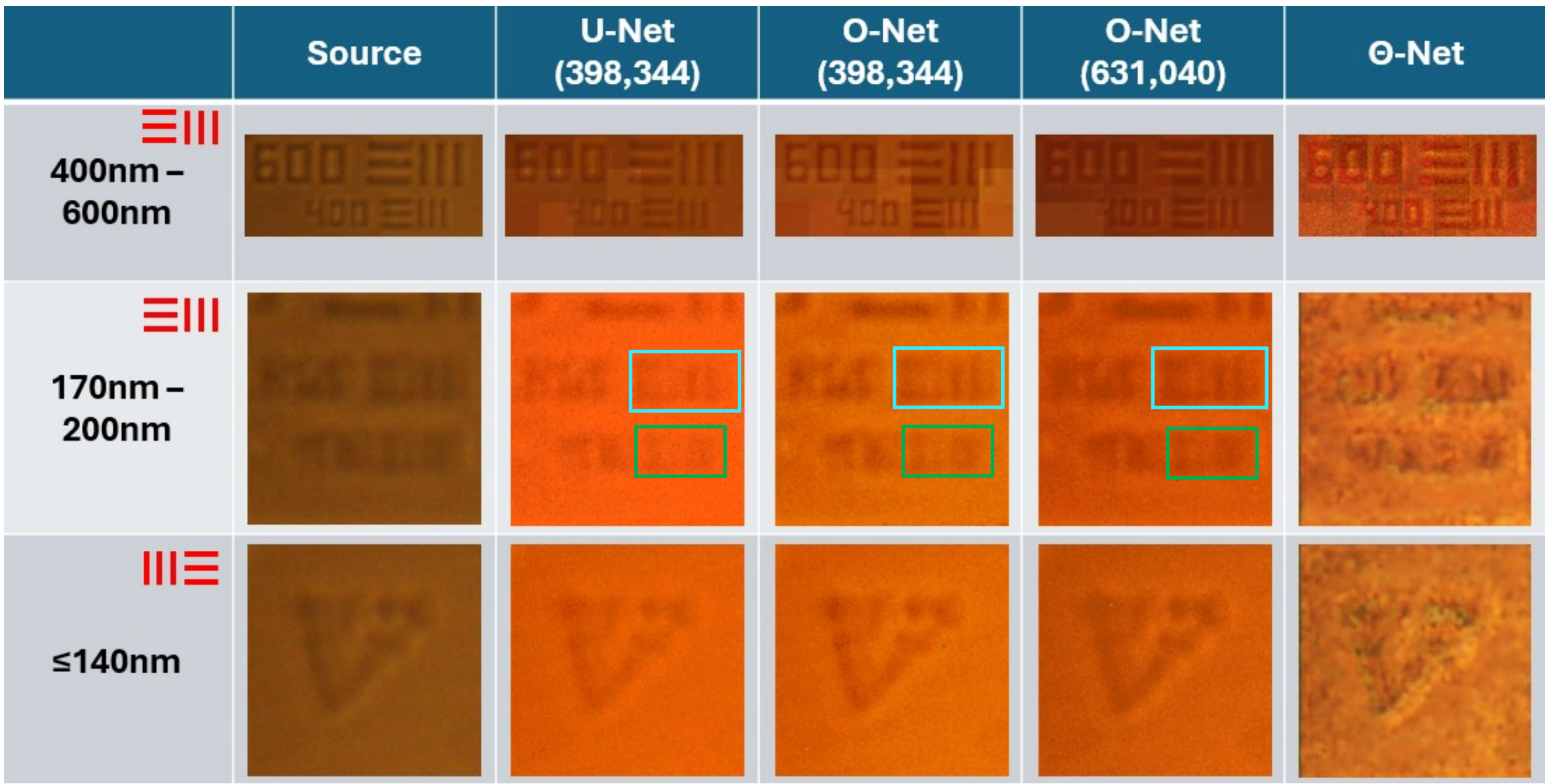

**Figure 3**: *In silico* SR DIC micrographs of the PMMA target, employing U-Net, O-Net & Θ-Net models trained over **101 epochs** (ep) (3944 iterations/epoch × 101 epochs = 398,344 iterations) and **160 epochs** (3944 iterations/epoch × 160 epochs = 631,040 iterations) (the 101-ep models employed a 5-layer architecture while the 160-ep models used a 7-layer architecture). Here, an intermediate magnification of **2.4X** was utilized for acquisition of the raw (**Source**) images, which correspond to the individual coloured boxed RoIs shown in Figure 2.4**D** previously (red for 400-600nm; blue for 170-200nm and green for ≤140nm features). Both the **Source** and model-generated images for features below 300nm were cropped as single 700px × 700px image tiles, while the 400-600nm RoI consists of stitched panoramas (7 × 3) of 300px × 300px image tiles (the brightness, contrast and sharpness of the generated images being Python-enhanced for easier comparison). From these images, one might observe that both the 101-ep and 160-ep O-Net-generated images demonstrate a clear delineation of the darker bars from the lighter-coloured spaces between them (especially for the 200nm bars, and perhaps less so for the 170nm bars), facilitating resolution of the said bars (with the 101-ep O-Net model outperforming the 160-ep O-Net model), while the U-Net generated image does not show such a clear distinct boundary (as depicted by the cyan & green rectangles). Additionally, the Θ-Net-generated images depict substantial noise for features ≤200nm, being far more sensitive to slight pixel variations (see Figure 4 for details).

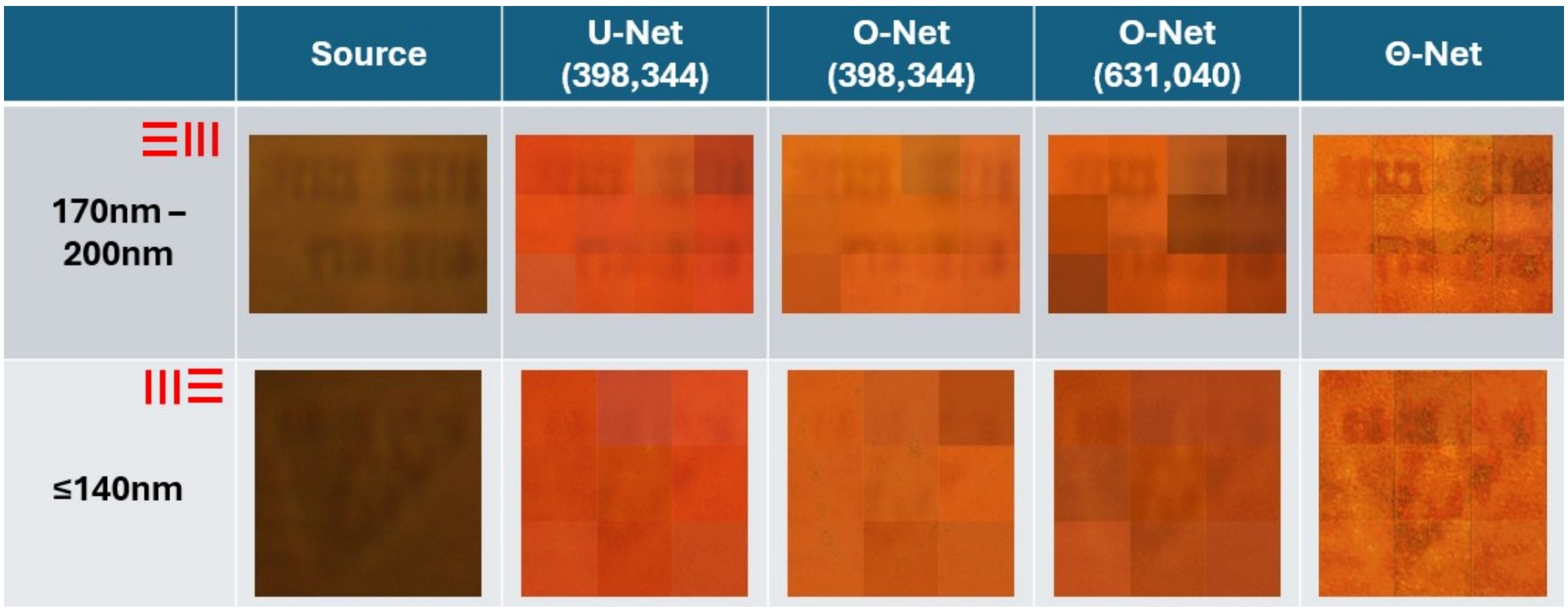

**Figure 4**: *In silico* SR DIC micrographs of the same custom-fabricated PMMA target depicted in Figure 3 previously. Here, an intermediate magnification of **3.75X** was used for acquiring the raw (**Source**) images, which were cropped into 256px × 256px image tiles. Intriguingly, the Θ-Net models seem to generate the most well-resolved images in the current context, surpassing that of U-Net and even O-Net (for both the 101-epoch & 160-epoch O-Net models). On closer inspection, it is observed that the **Source** images have a significantly reduced contrast to signal-to-noise ratio (SNR), making it rather challenging to resolve the individual bars in the target pattern. In such a situation, it is thus noteworthy to highlight that Θ-Net is well-placed to resolve these images, being more sensitive to minor fluctuations in pixel intensity (as mentioned previously).

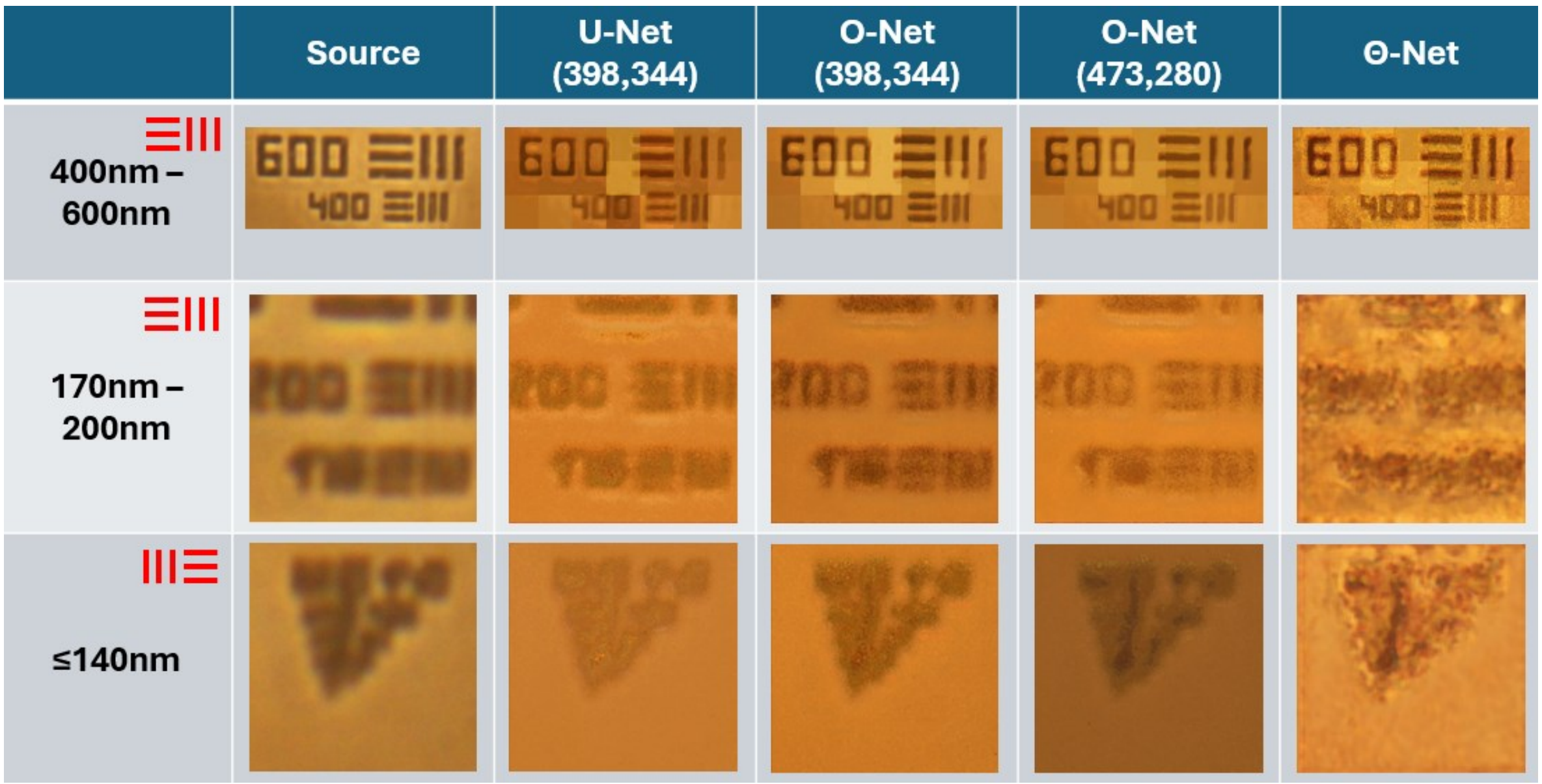

**Figure 5**: SR PCM micrographs of the custom-fabricated PMMA target, derived *in silico* via U-Net, O-Net and Θ-Net models being trained over **101 ep** and **120 ep** (3,944 iterations/epoch × 120 epochs = 473,280 iterations). As with Figure 3, a 2.4X intermediate magnification was employed to acquire the **Source** images. Also, as indicated in the caption

for Figure 3, the features between 400-600nm were cropped (and stitched) as 300px × 300px image tiles, while those features below 300nm were cropped as single 600px × 600px image tiles (RoIs were selected as demarcated via coloured boxes in Figure 2**C**). Due to the relatively high SNR of these PCM **Source** images, no enhancements were made to the generated images (unlike Figure 3 previously where the Source image SNR was relatively low). Here too, both the U-Net and O-Net models produced images showing a clear separation of the bars, with the best results gleaned from the 101-ep O-Net model. The Θ-Net models, however, depict significant image noise when super-resolving the PCM images, making them unsuitable for this use-case.

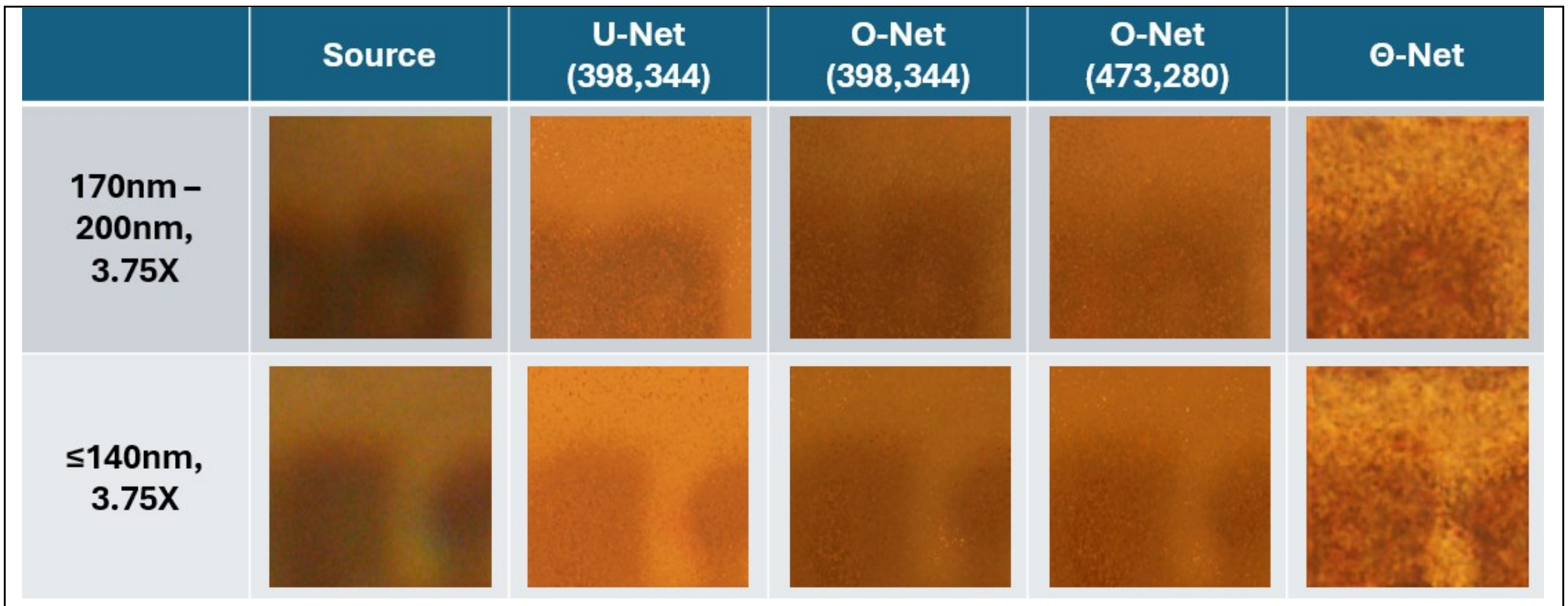

**Figure 6**: DNN-facilitated SR PCM micrographs of an image tile of the custom-fabricated PMMA target, employing U-Net, O-Net and Θ-Net models trained either over **101** or **120 epochs**. As with Figure 4, a **3.75X** intermediate magnification was used in capturing the raw (**Source**) images, prior to being cropped into 256px × 256px image tiles. As mentioned in the Figure 4 caption previously, the Θ-Net models have once again performed extremely well under this intermediate magnification, transcending both the U-Net and O-Net models when the **Source** image has a very low SNR.

Subsequently, the **Source** images for each of the boxed RoIs (depicted in Figures 3 and 5) were converted into grayscale in ImageJ 1.52n, prior to the intensity profile plotted for a linear RoI defined in the image (as described in the **Methodology**). The results of this analyses are depicted in Figures 7 and 8 as follows:

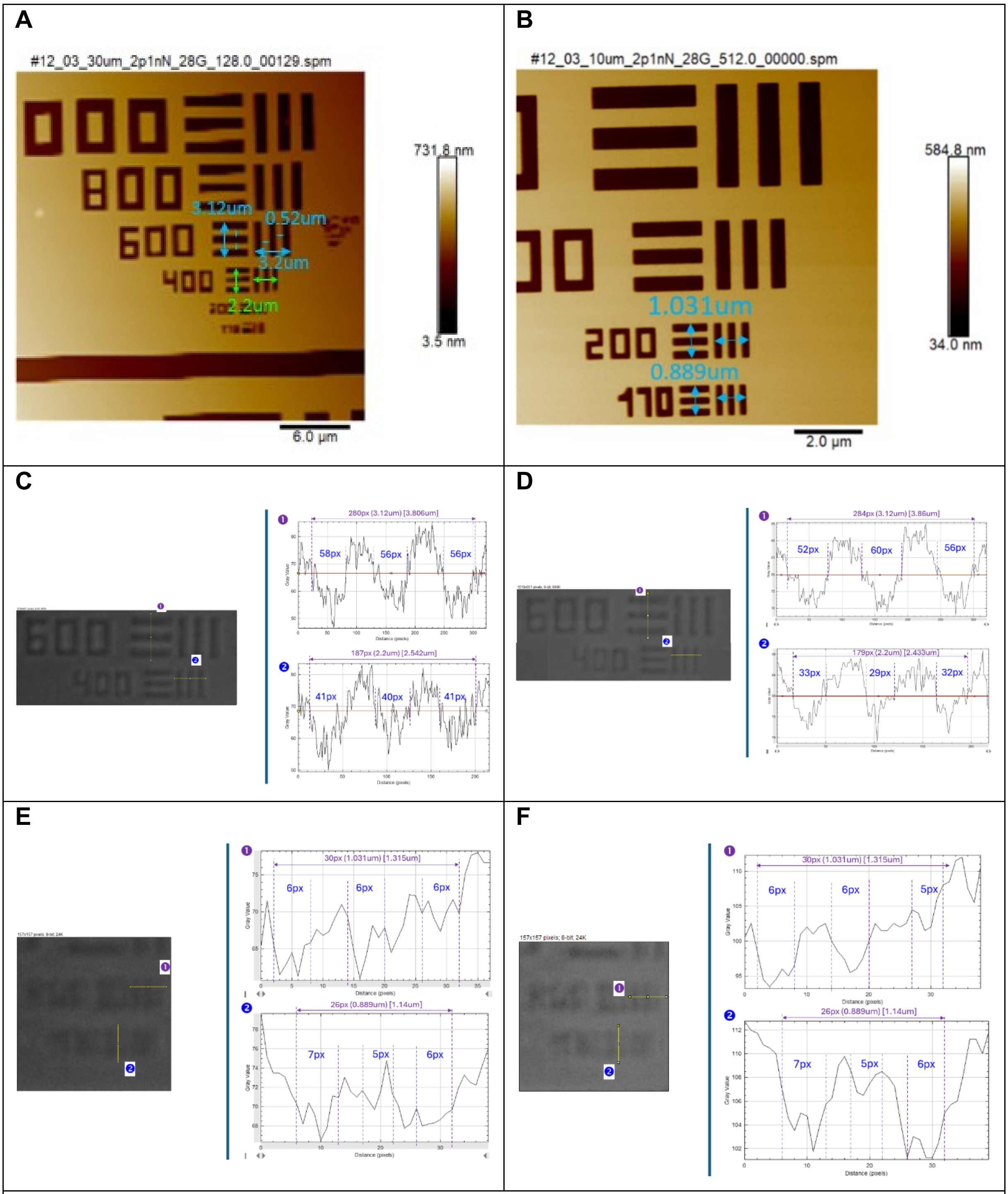

ꟷ **Figure 7**: **A** & **B** AFM-calibrated micrographs having their bar & space widths measured & indicated using the associated scale, with their respective grayscale-converted *in silico* SR DIC images (generated by the 101-ep O-Net model portrayed in Figure 3 previously) and intensity profile plots (**D** and **F**) as well as the corresponding **Source** images (**C** and **E**). In panels **C** to **F**, the measurements derived from panels **A** & **B** are indicated in parentheses, while the values in square brackets indicate the computed (*theoretical*) values as derived from the compounded magnification of the microscope's optical train. From the intensity profile plots in panel **F**, elevated levels of image noise (i.e. reduced SNR) make it highly challenging to isolate signal from noise using just a single linear threshold (as might commonly be done in image analysis). Due to this, separating the features (bars and spaces) is primarily achieved through visual comparison with the AFM micrograph (panel **B**), associating the fluctuations in pixel intensity with the actual grayscale values in the SR DIC images (panels **D** and **F**). From the mathematical construct of the Rayleigh criterion as presented in Equation (1) and the use of 3000K halogen light (corresponding to a peak emission of ~580nm), one may compute the diffraction-limited resolution as **308nm** (highlighted in Figure 2**E**), implying that the features in panel **F** are well below the diffraction limit. Panels **C** and **D**, however, do not depict these issues since their features are above the diffraction limit and are easily distinguished, allowing a single threshold (brown line) for feature isolation.

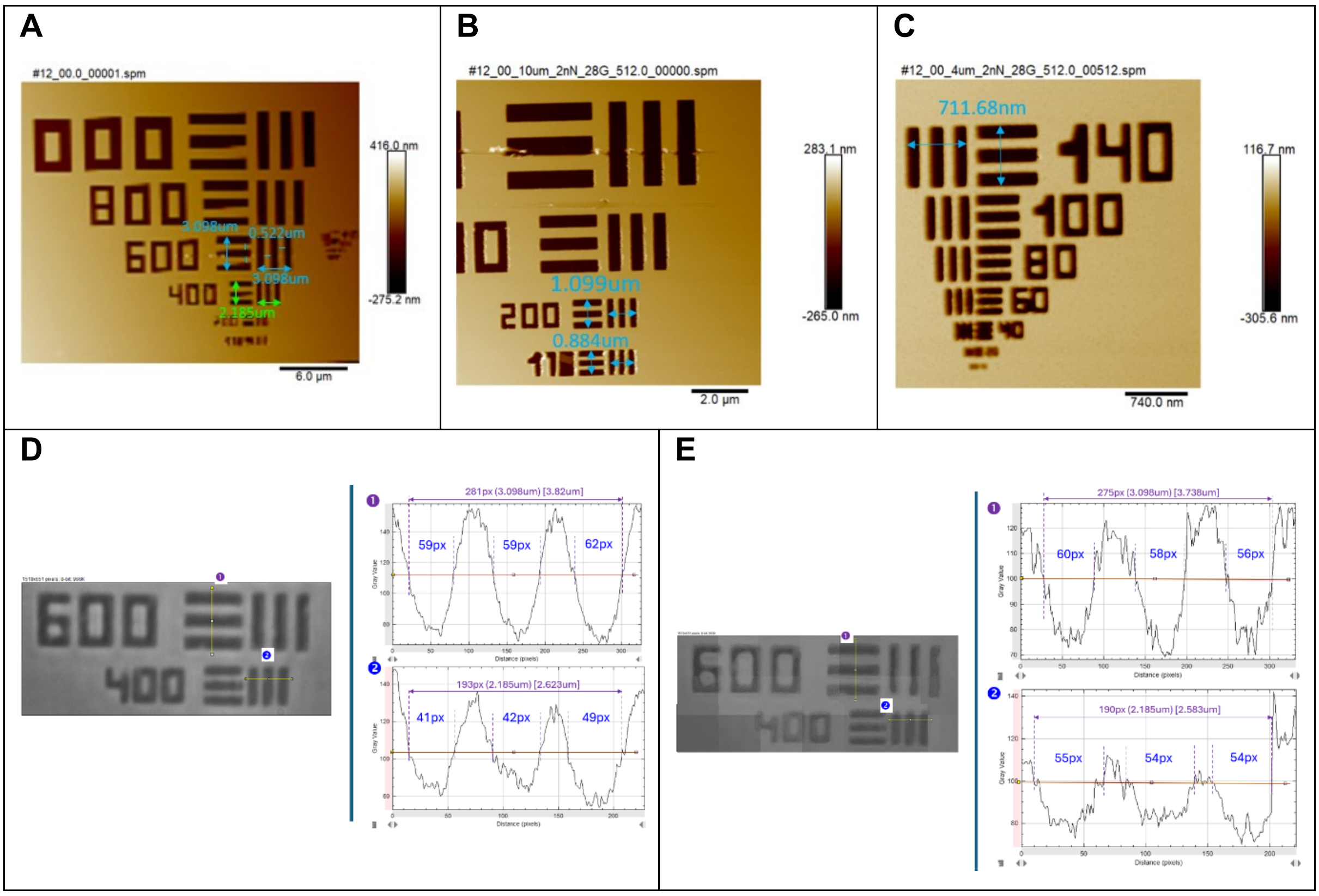

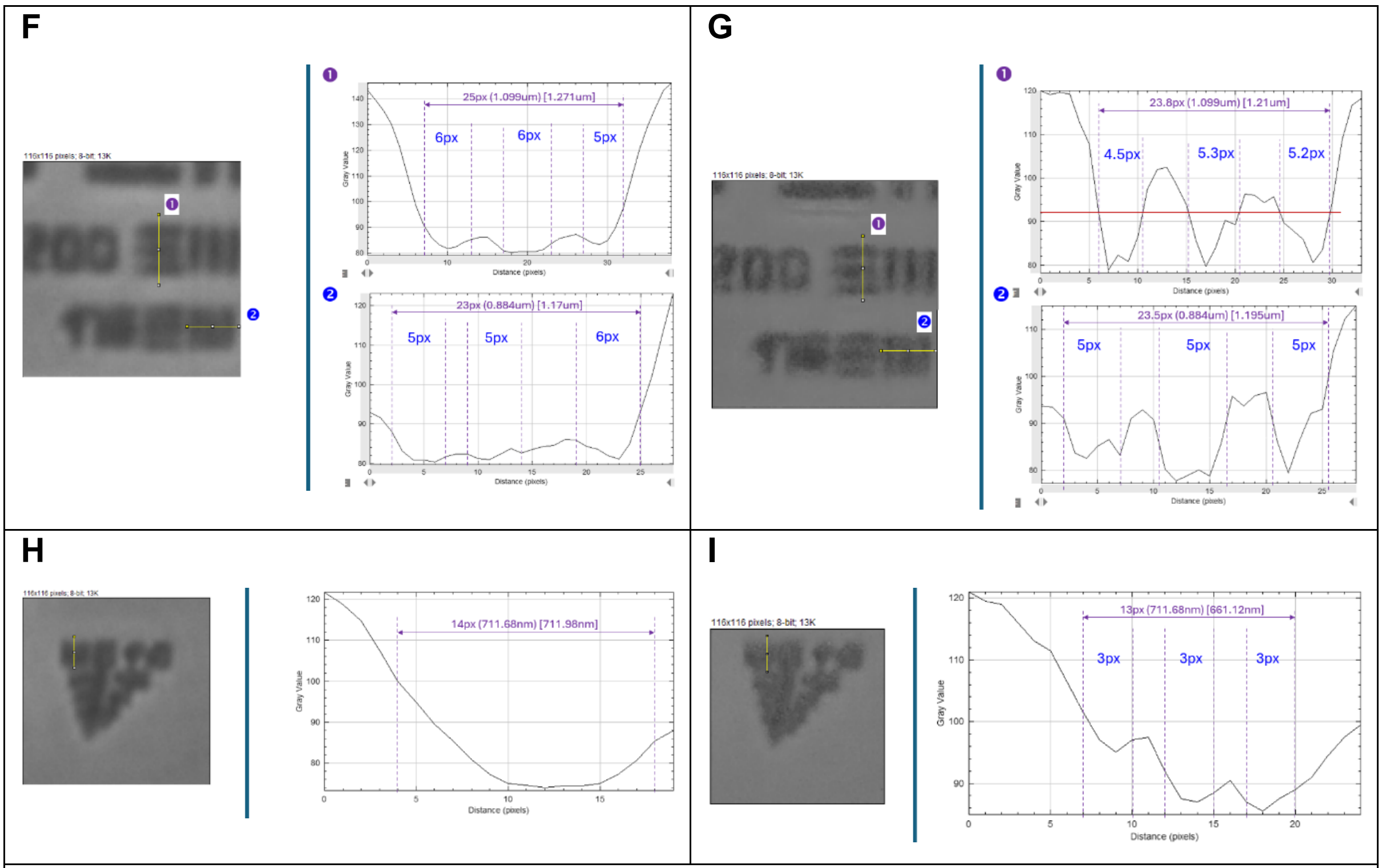

**Figure 8**: AFM-calibrated images with their respective measurements (panels **A**, **B** and **C**) coupled with their respective grayscale *in silico* SR PCM micrographs with the corresponding intensity profile plots generated via the 120-ep O-Net model (as shown in Figure 4) (panels **E**, **G** and **I**) and their input **Source** images (panels **D**, **F** and **H**). As in Figure 7 previously, the values in panels **D** to **I** indicate the measurements derived from panels **A** & **B** in parentheses, while the values in square brackets denote the theoretical values computed from the overall magnification of the microscope's optical train coupled with the camera pixel size. Notably, the discrepancies in the theoretical *vs* empirical values may be attributed to the fuzzy edges of the bars in the test target pattern, making it difficult to accurately define the boundary between a bar (trough) & its adjacent space (peak), which resonates with Figure 7 as well. For this reason, the theoretical values are not utilized in computing the resolution of the image. Also, higher levels of image noise in the SR PCM micrograph in panel **I** (for the ≤140nm features) makes it difficult to discriminate signal from noise, hence a similar approach adopted for Figure 7 was utilized in this context as well (i.e. the AFM micrographs were compared against the corresponding grayscale images to elucidate the boundaries segregating the bars from the spaces). The features in panel **H**, however, could not be resolved; an indication that these features lay well below the resolution limit. The 200nm features in panel **G** were easier to resolve though (attributable to the higher SNR of this RoI), whereas the 170nm features depicted increased image noise (similar to panel **I**), making it more challenging to isolate individual bars and spaces (a similar trend was noted for panel **F** as well, although the features were not as clearly

discernible as those in **G**, both visually and as evident from the intensity profile plot). As for panels **D** and **E**, the intensity profile plots do not manifest such issues, since their features are generally larger than 308nm, hence a single linear threshold may be used to demarcate the features (as in Figure 7). Additionally, the 200nm (and 170nm) features in panel **G** may also be considered to be smaller than the diffraction limit (308nm) despite being clearly visible in the O-Net model-generated image, demonstrating the capability of these models for SR PCM in the current context.

## Discussion

From Figures 3 and 5, it is noted that the O-Net models perform well in generating SR DIC and PCM images when image noise levels are low (i.e. high SNR), to the point of surpassing U-Net and Θ-Net models when paired alongside their corresponding AFM images. Notably, the Θ-Net models produce images with high levels of image noise and artifacts, portraying the Θ-Net model architecture's sensitivity to slight fluctuations in pixel intensity (triggering the formation of artefacts fuelled by noise). However, the converse is observed when evaluating images having a *reduced* SNR (i.e. high noise levels) such as is present under a higher intermediate magnification (3.75X) where the Θ-Net models have been reported to outperform both the U-Net & O-Net models in their generative SR potential instead (see Figures 4 and 6 for details). Another key point to be highlighted refers to how the SR PCM O-Net models seemingly resolved features down to 170nm (depicted in Figure 8**G**) while the SR DIC O-Net models were capable of resolving features down to 200nm (Figure 7**F**), thereby surpassing the theoretical Rayleigh diffraction limit of 308nm (see Figure 2**E** for details) and demonstrating the capability of these DNN models for use in computational label-free nanoscopical applications.

The inner workings of these assayed DNN models may be attributed partially to their ability to operate in the rear focal plane (Fourier domain) of the microscope's optical train, allowing the resolution of features from learnt convolution (*blurring*) as well as the corresponding transposed convolution (*deconvolution*) kernels. Image formation in the spatial domain (denoted as $g(x, y)$) may be expressed as the convolution of the *point spread function* (PSF) of the optical train $f(x, y)$ with the impulse response of the linear system $h(x, y)$, which can be characterized by Equation (2) below [23]:

$$g(x, y) = f(x, y) \circledast h(x, y) \text{ ----- (2)}$$

(where $\circledast$ represents convolution & $f(x, y)$ may be approximated by a Gaussian kernel).

It is also well-established in the literature ([23]) that the convolution operation in Equation (2) translates to *multiplication* in the Fourier domain, denoting an aspect which is often explored in *Fourier optics* [24]. When considering the operation of the models in the Fourier (*frequency*) domain, the trained DNN models are thus able to apply a *learnt* conjugate function to extend the Nyquist sampling limit, thereby allowing the resolution of features below the Rayleigh diffraction limit when extrapolated to the real (*spatial*) domain [23]. Interestingly, this does *not* require the models to be trained with frequency spectra acquired in Fourier space, although the perceived interoperability of the models across both the spatial & frequency domains in this respect would enable the resolution of features in both feature spaces. Moreover, this interoperability of the models would also account for their general resilience to *image noise* responsible for varying image SNR levels, since noise is often present as *random* high frequency components in Fourier space and thus would not be considered as a principal component to be integrated into the model weights during the training process. To illustrate this, one may consider the optical microscope as a 4*f* optical setup, where the real image coupled with its conjugate 2D frequency spectrum/Fourier transform (FT) in the rear focal plane of the microscope's objective lens may be isolated under differing approaches – Kohler illumination for the former *vs* critical illumination with conoscopic examination for the latter – as is described in Figures 9 & 10 below:

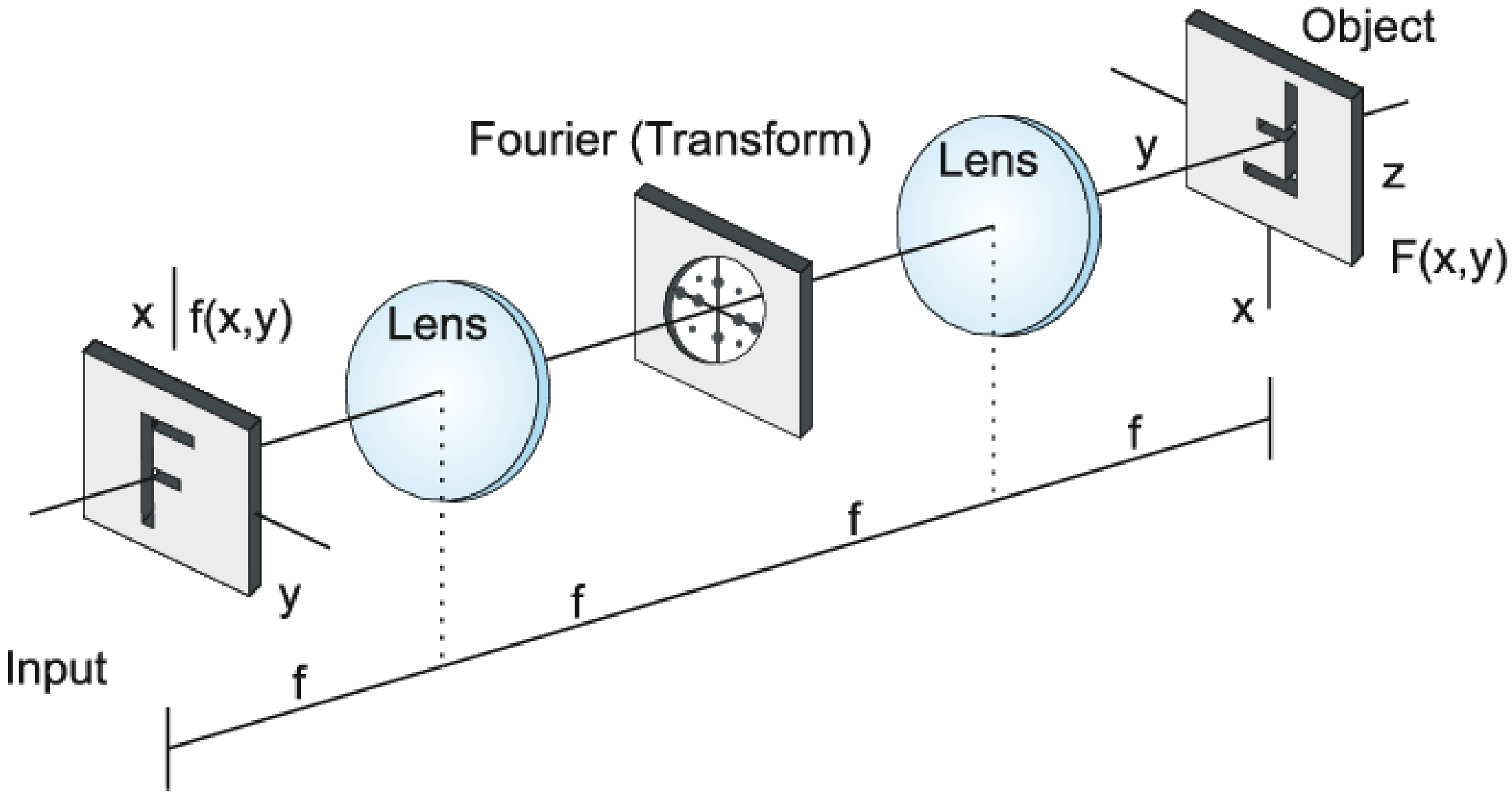

**Figure 9**: Transmission of light rays through two lenses arranged in a 4*f* optical setup; the image being formed at the image plane (4*f*) while the frequency spectrum is observed at the Fourier plane between the two lenses (2*f*). Higher spatial frequencies (contributed by smaller structures) are generally found at the periphery of the frequency spectrum, while the central maximum corresponds to the lower spatial frequencies (larger structural features) [8]. The PSF of the optical system would thus also have its FT projected onto the rear focal plane of the microscope's objective lens – an aspect known as the *optical transfer function* (OTF) [8]. Figure Source: [26].

Further exemplification of the light microscope as a classic 4*f* optical instrument enabling it to receive signals in both the spatial & frequency (Fourier) domains may be depicted in Figure 10 below, where we have attempted to image a specimen using conventional brightfield microscopy & obtain its corresponding Fourier field (acquired under critical illumination):

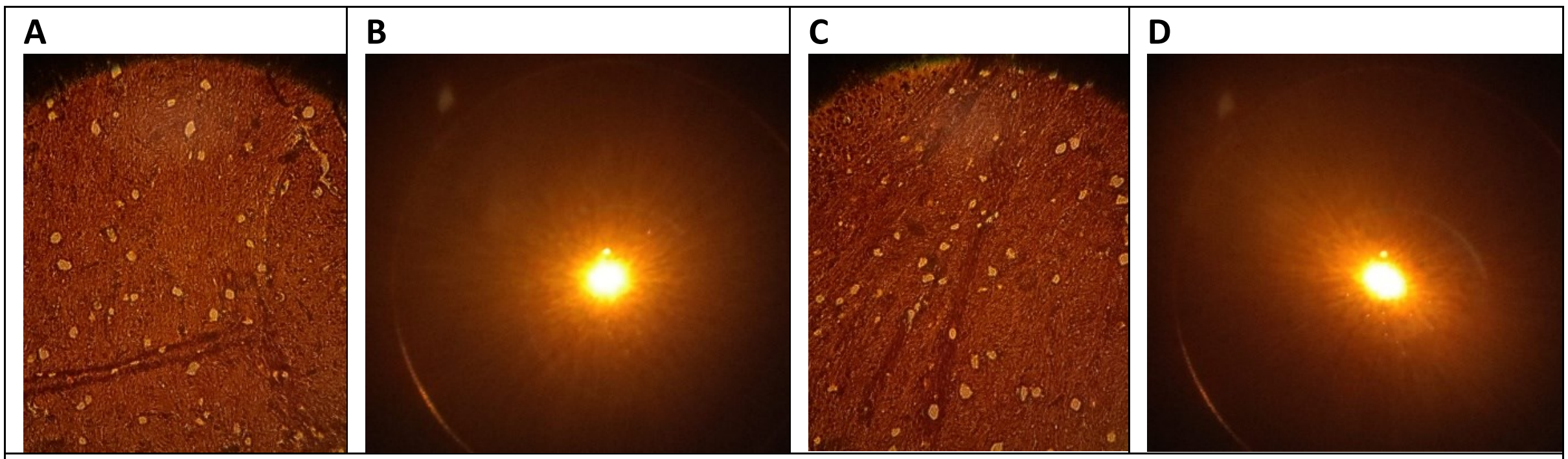

ɵ **Figure 10:** Panels **A** & **C** depict 2 separate RoIs imaged from the same specimen (*rabbit cerebrum section*, *silver stained*). Panels **B** & **D** portray the complex Fourier image field of this same tissue sample with radiating spokes depicting the spatial frequencies observed in the rear focal plane of the microscope's objective lens for *each* of the RoIs shown in panels **A** & **C** respectively. Notably, each point in this pattern (represented by an infinitesimally small point light source) describes a sinusoidal function $I(\boldsymbol{r}) = A\sin(\boldsymbol{k}_{(x,y)} \cdot \boldsymbol{r} + \varphi_0)$ of a specific amplitude *A* and phase $\varphi_0$, as well as 2D frequency & orientation vector $\boldsymbol{k}_{(x,\ y)}$ (direction of propagation of the wave) [27]. Notice the differences in the observed Fourier image fields – features with high spatial frequencies are localized at the periphery of the spectra, while lower frequency features are often clustered in the centroid region. Also as highlighted in [26], the Fourier image fields (**B** & **D**) resemble their respective object fields (**A** & **C**) but are convolved with the Fraunhofer diffraction pattern of the pupil, i.e. a Fourier transform applied to the pupil function.

*Potential Advantages and Limitations*

Further evaluation of the results presented in this study reveals a minimum lateral resolution of 200nm (for SR DIC) and 170nm (for SR PCM) for the O-Net-generated images, implying a ~1.54-fold and a 1.81-fold increase in resolution for SR DIC and SR PCM respectively (when considering the Rayleigh diffraction limit of 308nm). Nonetheless, the 170nm features in SR DIC (Figure 7**F**) and 140nm features in SR PCM (Figure 8**I**) are very slightly resolvable, clearly demonstrating the potential of the underlying O-Net models in further overcoming the Rayleigh diffraction barrier to attain image SR. In this light, future potential improvements that could be implemented in the proposed model architecture might involve determining the exact image shear introduced by the Nomarski prisms for the SR DIC images and using this to compute the shear angle ε (in radians), which can then be used to derive the optical path difference (bias) $\Gamma$ and retardance δ (in degrees) in accordance with the following equation (adapted from [25]):

$$\varepsilon = \frac{d\Gamma}{dx}\text{ , where } \Gamma = \frac{\delta}{360^{\circ}}\lambda \ (\lambda \text{ – the irradiating wavelength}) \text{ ---------- } (3)$$

The term ε can then be integrated into the loss function used for training the models, thereby minimizing potential shear bias introduced through translating the objective prism while enhancing model resilience to pseudo-relief fluctuations.

It would also be prudent to mention at this juncture that there is presently *no* feasible image SR metric which can be used to objectively quantify the performance of a DNN model in super-resolving non-fluorescent micrographs for which there is *no* precedent/ground truth image to compare the generated image against. The Fourier Ring Correlation (FRC) metric ( [26], [27] & [28]), which is often cited as an objective & quantitative measure of image resolution, requires two independent images of the same RoI for comparative assessment & can potentially only be used on fluorescent or darkfield images (where the background is totally dark). Due to this, one possible approach to quantify image resolution for non-fluorescent micrographs (as in this context) would involve the use of an intensity profile plot across a defined RoI in the image, and the establishment of a threshold for the relative intensities corresponding to the individual bars/spaces. However, this method will still face a caveat when attempting to resolve structures well below the diffraction limit, such as ≤140nm in this study (referring to Figure 8**H** for details). Under such situations, the boundary between the bars and the spaces becomes increasingly blurry as pixel intensity varies in a somewhat gradual manner, making it challenging to accurately infer the specific locations corresponding to either the bars or spaces. This dilemma is further exacerbated through the inclusion of multiple other factors, such as contrast of the input (**Source**) image influenced by varying illumination intensities, camera exposure levels and image gain.

On a separate note, it will be essential to note that the developed models in the current context are also intended for use with the current microscope/camera setup [i.e. a Leica DM4000M microscope (Leica Microsystems GmbH) coupled with a RisingCam® (RisingTech) E3ISPM12000KPA CMOS camera]. Should a different microscope/camera be used, this would undoubtedly alter the point spread function (PSF) of the optical train, thereby potentially necessitating a retraining of the models.

## Conclusion

In the present study, the image SR performance of models trained on two of our developed DNN architectures (O-Net [20] and Θ-Net [21]) was evaluated against U-Net [29] (a model architecture widely-utilized in the industry for computer vision and image sensing applications). To ensure a fair and unbiased assessment, a calibrated test target (employing the USAF1951 standard of equally-spaced bars) was utilized as a benchmark of image resolution, the said target being verified under AFM. From the presented results, it is evident that both the O-Net and Θ-Net models performed relatively well, albeit under very different conditions – O-Net excelled where the image SNR was high, while Θ-Net outperformed both O-Net and U-Net models under conditions of low image SNR. It would thus be noteworthy to highlight that one's choice between the use of O-Net or Θ-Net models for image SR when performing *in silico* phase-modulated nanoscopy would depend very much on the image SNR of the acquired **Source** image, marking important applications of each of these models in the industry, particularly in fields such as biomedical research, materials sciences & in quality control, to facilitate the identification of sub-microscale defects in computing chips, or to assay phasic variations indicative of metabolic processes *in vivo*.

In retrospect, rapid advancements in graphical processing unit (GPU) hardware developments coupled with the relatively low cost of computing resources are fuelling the development of more extensive & elaborate DNN model architectures, with additional layers & operations included (where feasible). This may indicate the potential for the current O-Net & Θ-Net DNN architectures in also incorporating such additional functions, although other niche applications (such as in metrology, non-destructive testing and quality control) might necessitate a re-training of the said models to attain true image SR, especially where a different optical setup is being employed.

## Data Availability Statement

The data that support the findings of this study are not openly available due to reasons of sensitivity and are only available from the corresponding author upon reasonable request.